\documentclass[letterpaper, 10 pt, conference]{ieeeconf}  

\IEEEoverridecommandlockouts                              

\usepackage{microtype}
\usepackage{amsmath}
\usepackage{amssymb}
\usepackage{amsfonts}

\usepackage{cite}

\usepackage[pdftex]{graphicx}
\DeclareGraphicsExtensions{.pdf,.jpeg,.png}

\usepackage{multirow}
\usepackage{booktabs}
\usepackage{tikz}
\usetikzlibrary{positioning}

\usepackage[ruled,vlined,linesnumbered]{algorithm2e}
\usepackage{algcompatible}
\SetAlCapNameFnt{\small}
\SetAlCapFnt{\small}
\SetAlCapSty{textsc}
\SetAlCapNameSty{textsc}

\newcommand{\minus}{\scalebox{0.5}[1.0]{$-$}}

\usepackage[caption=false,font=footnotesize]{subfig}

\usepackage{url}

\newcommand{\etal}{\textit{et al. }}

\usepackage{float}
\usepackage{xargs}

\usepackage[hidelinks]{hyperref}
\usepackage{cleveref}
\Crefname{figure}{Fig.}{Figs}

\title{\LARGE \bf
Vehicle Behavior Prediction and Generalization Using \\ Imbalanced Learning Techniques
}

\author{Theodor Westny, Erik Frisk, and Björn Olofsson 
 \thanks{This research was partially supported by the Strategic Reseach Area at
 	Linköping-Lund in Information Technology (ELLIIT), and partially
 	supported by the Wallenberg AI, Autonomous Systems and Software
 	Program (WASP) funded by the Knut and Alice Wallenberg Foundation.}
 \thanks{The authors are with the Division of Vehicular Systems, Department of Electrical Engineering,
 	Linköping University, Linköping, Sweden. \scriptsize{\{\texttt{theodor.westny, erik.frisk, bjorn.olofsson}\}@\texttt{liu.se}}}
}

\begin{document}

\maketitle
\thispagestyle{empty}
\pagestyle{empty}

\begin{abstract}
	The use of learning-based methods for vehicle behavior prediction is a promising research topic.
	However, many publicly available data sets suffer from class distribution skews which limits learning performance if not addressed.
	This paper proposes an interaction-aware prediction model consisting of an LSTM autoencoder and SVM classifier.
	Additionally, an imbalanced learning technique, the \emph{multiclass balancing ensemble} is proposed.
	Evaluations show that the method enhances model performance, resulting in improved classification accuracy.
	Good generalization properties of learned models are important and therefore a generalization study is done where models are evaluated on unseen traffic data with dissimilar traffic behavior stemming from different road configurations. 
	This is realized by using two distinct highway traffic recordings, the publicly available NGSIM US-101 and I80 data sets.
	Moreover, methods for encoding structural and static features into the learning process for improved generalization are evaluated.
	The resulting methods show substantial improvements in classification as well as generalization performance.
\end{abstract}

\section{Introduction}
Autonomous driving research has seen a tremendous interest during
recent years and developing fully self-driving cars has engaged 
academia and industry
alike. 
Benefits of adopting autonomous vehicles on public roads
are forecast to be numerous and there is particular emphasis on the
 enhancement of road safety. To develop safe and efficient
autonomy, vehicles need to be socially compliant and exhibit
conventional behavior.
This requires that the vehicle has the ability to anticipate the motion and intention of surrounding vehicles.

\begin{figure}[!t]
  \centering
  \includegraphics[width=3in]{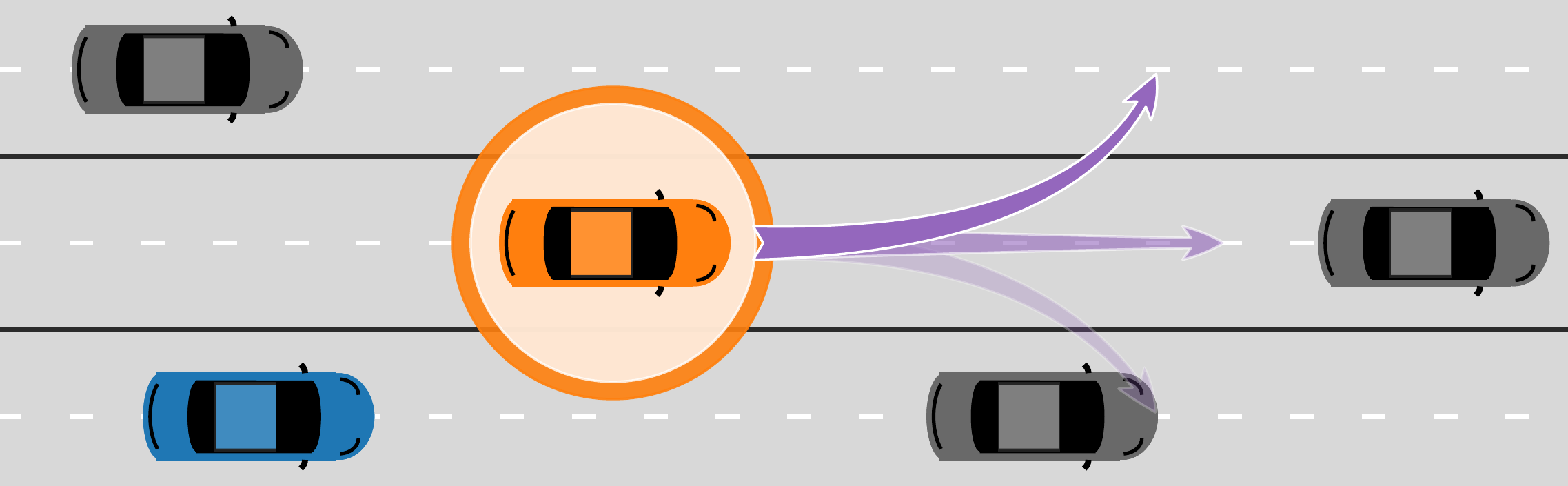}
  \caption{The prediction task, with the ego vehicle shown in blue and the target vehicle in orange. 
  The arrows correspond to the considered possible actions.}
  \label{fig:lane_change}
  \vspace{-0.12in}
\end{figure}

Vehicle behavior prediction may be categorized into two general classes, \textit{motion prediction} and \textit{intention prediction}.
For the motion prediction problem, the task is to predict the future motion/trajectory of a target vehicle (TV).
Intention prediction, however, may vary depending on the studied scenario but the task is almost always the same---to infer the probability for a set of predefined behaviors.
This includes inferring \textit{route patterns}, e.g., predicting the planned action of a TV when approaching an intersection.
The prediction of \textit{social patterns} also falls under the category of intention prediction, e.g., quantifying the cooperativeness in a forced merging scenario.
In this work, we are specifically interested in predicting maneuver
intention in highway scenarios where the proposed method should infer
either a lane-changing or lane-keeping action over an extended
horizon, schematically illustrated in \Cref{fig:lane_change}.

The prediction tasks may be further categorized depending on the
available information included from observations.
Notably, \textit{interaction-aware models} consider
inter-vehicle dependencies using information on surrounding vehicles
(SVs) \cite{lefevre2014survey}. Interaction-aware
models are hypothesized to improve predictions since SVs limit the availability of some actions.

As a results of the inherent multi-modality of these particular problems, deep learning has become a prominent tool for vehicle behavior prediction \cite{mozaffari2020deep}.
The benefit of using learning-based methods is credited to their proven ability to find structure in complex data.
The use of deep networks for behavior prediction has provided an advantage for interaction-aware modeling by the inclusion of information on SVs without explicit rules for inter-vehicle dependencies \cite{deo2018convolutional, messaoud2020attention, hu2018probabilistic, ding2019predicting}.
Instead, it is included within the task of these models to learn the connections between interacting traffic agents.

Although promising for behavior modeling, learning methods require certain preconditions to learn properly.
In particular, many methods struggle, unless properly addressed, with achieving prescribed performance when faced with an imbalanced data distribution \cite{he2013imbalanced}.
Uneven class distributions are typical in many real-world applications
and this is also the case for the investigated highway recordings in
this work, where the number of lane-keeping instances severely
outnumber the number of lane-changes. 
A common method found in related works
(see \Cref{sec:related_work}) is to remove majority class
instances until an even balance is obtained. Although an
effective technique to reach balance, this
approach discards potentially useful examples.

This paper includes several topics of interest for vehicle behavior prediction. 
The constant throughout the work is the investigation and development of imbalanced learning techniques for behavior prediction.
Secondly, we investigate the representativeness of information in two different highway trajectory data sets and
 how generalization of learned models carries over to different traffic scenarios together with the proposed methods for improvement.

\subsection{Related Work}
\label{sec:related_work}
Vehicle behavior prediction is currently a prominent research topic.
A survey on vehicle behavior prediction methods has been presented by Lefèvre \etal \cite{lefevre2014survey}.
A more recent survey by Mozaffari \etal focuses on learning-based approaches \cite{mozaffari2020deep}, which is the topic of this section.
Previous work on driver/vehicle behavior prediction proposes various solutions,
distinguished by model choice, including using Support Vector Machines (SVMs) \cite{kumar2013learning, woo2017lane},
 Hidden Markov Models (HMMs) \cite{streubel2014prediction, deo2018would}, Multilayer Perceptron (MLP) \cite{yoon2016multilayer},
  Recurrent Neural Networks (RNNs) \cite{phillips2017generalizable, zyner2017long, deo2018convolutional, messaoud2020attention, ding2019predicting, dang2017time, messaoud2020trajectory, zhao2019multi},
   and Convolutional Neural Networks (CNNs) \cite{lee2017convolution, casas2018intentnet, cui2019multimodal, deo2018convolutional}.
The model choice is often decided by the type of available data as well as the prediction task.

For the motion prediction task, the output of the model is a sequence with length corresponding to the prediction horizon,
and so, a popular method for such tasks is to use recurrent neural networks and in particular Long Short-Term Memory (LSTM) RNNs.
In \cite{altche2017lstm}, Altché \etal proposed an LSTM-based network for highway trajectory prediction.
Historic TV and SV features were fed into an LSTM encoder and passed into a fully-connected layer to infer the future trajectory.
Zyner \etal \cite{zyner2019naturalistic} proposed an encoder--decoder network for path prediction in urban environments.
The model was further equipped with a Mixture Density Network (MDN) to handle the multi-modal nature of the trajectory data.
In \cite{deo2018convolutional}, Deo \etal proposed an interaction-aware method for vehicle trajectory prediction in highway scenarios.
The authors proposed an LSTM encoder--decoder model together with a \textit{social pooling} \cite{alahi2016social} layer for SV spatial configuration encoding.
The approach is extended by Zhao \etal in \cite{zhao2019multi} to also incorporate a scene encoding that contains the current static context.
In \cite{messaoud2020attention}, Messaoud \etal proposed the use of an \textit{attention} mechanism \cite{bahdanau2016neural} to learn the explicit importance of SVs in an encoder--decoder model for trajectory prediction.
In \cite{messaoud2020trajectory}, they extend their method to use a joint agent-map representation as input.

The task of motion prediction is tightly coupled with the intention prediction task.
This is utilized in the work by Woo \etal \cite{woo2017lane} where they proposed a two-stage approach: first they predict the trajectory, which is then used to infer lane-change behavior.
The reverse approach is investigated by Xin \etal in \cite{xin2018intention} where a dual LSTM network approach was employed.
The first part of the network was tasked with classifying the vehicle's intention, which is then fed into the next part of the network that should predict the future trajectory.
In \cite{hu2018probabilistic}, Hu \etal proposed a Gaussian Mixture Model (GMM) for generalized intention and motion prediction by using semantically defined behaviors.
Instead of predefining a discrete set of actions, they identify spaces within the environment that the TV could occupy and thus predict the intention and subsequently---vehicle motion by their probability to occupy an insertion area in the future.
Finally, there are also prior works that are mainly concerned with the intention prediction task, for example \cite{phillips2017generalizable, zyner2017long, ding2019predicting, lee2017convolution} to name a few.
Although input data may vary, these are considered the works which are most closely related to this work in terms of problem formulation.

\subsection{Contributions}
The following are our main contributions:
\begin{enumerate}
  \item \textbf{Behavior modeling} --- We present an interaction-aware
    vehicle-intention prediction model consisting of an autoencoder
    coupled with a support vector machine classifier,
    \textit{AE-SVM}. Classification performance is evaluated and classifier
    uncertainty is quantified for different time-to-lane-change.
  \item \textbf{Imbalanced learning} --- The \emph{Multiclass Balancing Ensemble} (\emph{MCBE}) technique is presented for the improvement of vehicle behavior prediction using imbalanced data sets.
  \item \textbf{Generalization study} --- We assess the generalization ability of learning-based vehicle behavior prediction models between similar,
   although distinct, highway data sets with different structural
   properties. The effects of including static features are also explored.
\end{enumerate}

Implementations are available online\footnote{ \url{https://github.com/westny/imb-behavior-prediction}}.


\section{Problem formulation}
The prediction problem is formulated as estimating the probability distribution of a discrete set of predefined actions,
lane-keeping (LK), lane-change left (LCL), and lane-change right (LCR) over an extended prediction horizon.
Observations used for training and inference are multivariate time series, composed of trajectory histories and relative positional information
from real recorded highway traffic data.

Consider an intention prediction model $\mathcal{M}$.
Given an observation $\mathbf{x}^{(i)} \in \mathbb{X} = \bigcup_{i=0}^n \mathbf{x}^{(i)}$ from the set of observations $\mathbb{X}$, the model $\mathcal{M}$ should output the predicted maneuver class $\hat{\mathbf{y}}^{(i)} = \mathcal{M}(\mathbf{x}^{(i)})$ as well as its confidence.
Thus,  $\mathcal{M}$ should output the conditional distribution $\text{Pr}(\mathbb{Y} |\mathbb{X})$ where $\mathbb{Y}$ is the finite set of actions $\mathbb{Y} = \{\text{LCL, LK, LCR}\}$.
The problem is cast as a classification task where the model is trained by its ability to correctly label observations.

Although the models are trained and evaluated based on their ability to correctly classify observations,
in practice it is in addition interesting to infer the \emph{probability} of a given action.
Therefore, the models are also evaluated with regards to their prediction confidence,
and in particular when faced with LC maneuvers and how their confidence change over time when nearing the time-to-lane-change (TTLC) instant.


\section{Proposed Method}
\label{sec:proposed_method}

\subsection{Data Set}
For model training and evaluation, publicly available data sets provided by the Federal Highway Administration (FHWA) are used.
The data sets contain recorded highway traffic data captured at 10 Hz from US-101 \cite{colyar2007us101} and Interstate 80 (I80) \cite{halkias2006i80}.
The data sets have been studied extensively in research of microscopic traffic behavior 
and more recently seen frequent use in learning-based prediction studies \cite{deo2018convolutional, ding2019predicting, messaoud2020attention, xin2018intention, woo2017lane, altche2017lstm, hu2018probabilistic}.
Both data sets provide 45 min worth of recordings divided into three 15-minute subsets simulating mildly, moderately, and heavily congested traffic.
The NGSIM highway data sets are consistently used in many studies, but have been reported to suffer from considerable tracking noise \cite{montanino2013making}.
Notably, the authors in \cite{altche2017lstm} proposed to use a smoothing (Savitzky-Golay) filter on the trajectories before inputting into their network.
In this work, noise filtering is included within the learning task and motivated by the model choice (see \Cref{sec:sys_arch}).

The segments of the two highways where the traffic recordings were captured share many structural similarities.
Although the recorded segment of the I80 contains a merging on-ramp, 
the US-101 segment has an auxiliary lane that accommodates for both entering and exiting the highway.
Because of its more dynamic nature, the auxiliary lane comprises a large amount of LCs and is one of the reasons the US-101 contains more LCRs than the I80.
The differences between the two data sets affect the traffic behavior which is investigated in the generalization study (see \Cref{sec:gen_study}).

\subsubsection{Preprocessing}
\label{sec:data_extraction}
Observations for training and inference contain 40 frames (4 s) of feature history.
Upon extraction, observations are downsampled by a factor of 2 for memory efficiency and reduced model complexity.
The length of the prediction window is set to 4 s and labeling an observation as an LC requires that the vehicle has changed lanes at the end of the prediction window.
The observations are then stored together with the corresponding TTLC value, which for LK instances is set to 6 s.
Knowing beforehand of the abundance of LK instances in the data sets, different strategies are employed depending on the identified maneuver.
LC instances are collected more generously with some overlap while LK are collected to a lesser extent. 
This is controlled by varying the sliding window step size during extraction.
Even so, the imbalance is still prominent.
The number of LKs is halved in a random fashion yielding the distribution of maneuvers shown in \Cref{tab:data_imbalance}.

\begin{table}[!h]
	\caption{Distribution of maneuvers}
	\label{tab:data_imbalance}
	\centering
	\begin{tabular}{c c c c} 
		\toprule
		Data set & LCL & LCR & LK \\
		\midrule
		\emph{US-101} & 30281 \emph{(3.3\%)} & 19034 \emph{(2.1\%)} & 874245 \emph{(94.6\%)} \\ 
		\emph{I80} & 48284 \emph{(5.1\%)} & 10938 \emph{(1.2\%)} & 882017 \emph{(93.7\%)} \\ 
		\midrule
		Total used & 78565 \emph{(7.9\%)} & 29972 \emph{(3.0\%)} & 878131 \emph{(88.1\%)} \\
		\bottomrule
	\end{tabular}
\end{table}

\subsubsection{Generalization Across Data Sets}
The generalization properties of behavior models are essential in practice since it is expected that performances carry over to different road networks.
The availability of two different highway trajectory data sets offers an opportunity for investigation of how learned models generalize across different traffic scenarios.

Consider two similar but disjoint data sets, $\mathcal{D}_1$ and $\mathcal{D}_2$.
The model $\mathcal{M}$ is then trained by minimizing the loss function
\begin{equation}
    \mathcal{L}(\mathbf{y}, \mathcal{M}(\mathbf{x})), \quad \forall (\mathbf{y},
    \mathbf{x}) \in \mathcal{D}_1. 
\end{equation}
Model performance is then evaluated by its accuracy on $\mathcal{D}_2$.
In this work, $\mathcal{D}_1$ and $\mathcal{D}_2$ represent the two vehicle trajectory recordings from US-101 and I80.

\subsection{Input Features}
\label{sec:input_features}
The used data sets offer a collection of vehicle trajectories and structural information from recorded highway data.
It is here assumed that this data may be gathered by the ego vehicle on-board sensors. 
The features used make up the set of observations $\mathbb{X}$ and may be divided into TV features, interaction (SV) features, and static features.

\subsubsection{Target Vehicle Features}
The features pertaining to the TV are summarized in \Cref{tab:tv_feat}.
Two features are used to represent lateral position.
The first, denoted with $y$, represents the evolution of the vehicle's lateral coordinates with respect to its initial position $y_0$ for the recorded maneuver.
The second feature $d_y$ refers to the TV's lateral deviation from the recorded centerline and is bounded to the interval $[-1, 1]$ 
according to 
\begin{equation}
    d_y = 2\frac{y_l - l_{y,l}}{l_w}-1,
\end{equation}
where $y_l \triangleq y+y_0$ refers to the local coordinate of the vehicle in the recorded data frame,
 $l_{y,l}$ is the lateral coordinate of the left lane divider for the current lane,
 and $l_w$ is the lane width.
Feature $d_y$ may be very informative for LC detection, but without knowledge of $y$, the feature by itself could be misleading.
Consider a vehicle that has just changed lanes during an observation; by only studying $d_y$ it is difficult to distinguish between an immediate return to its initial lane or continued motion away from it.
Such a distinction is crucial in a real-world application since the former example may indicate an emergency maneuver.
Finally, although the investigated data sets are highway segments with low curvature, the feature $d_y$ is adaptable to road geometry, which is not the case for $y$.

\begin{table}[!h]
	\caption{Target vehicle feature selection}
	\label{tab:tv_feat}
	\centering
	\begin{tabular}{c l c}
		\toprule
		Feature & Description & Unit\\
		\midrule
		$y$ & Lateral coordinate with respect to $y_0$ & m\\ 
		$d_y$ & Lateral deviation from the current lane centerline & -\\
		$v_y$ & Instantaneous lateral velocity & m/s\\
		$v_x$ & Instantaneous longitudinal velocity & m/s \\
		\bottomrule
	\end{tabular}
\end{table}

\begin{figure}[!t]
    \centering
    \includegraphics[width=3in]{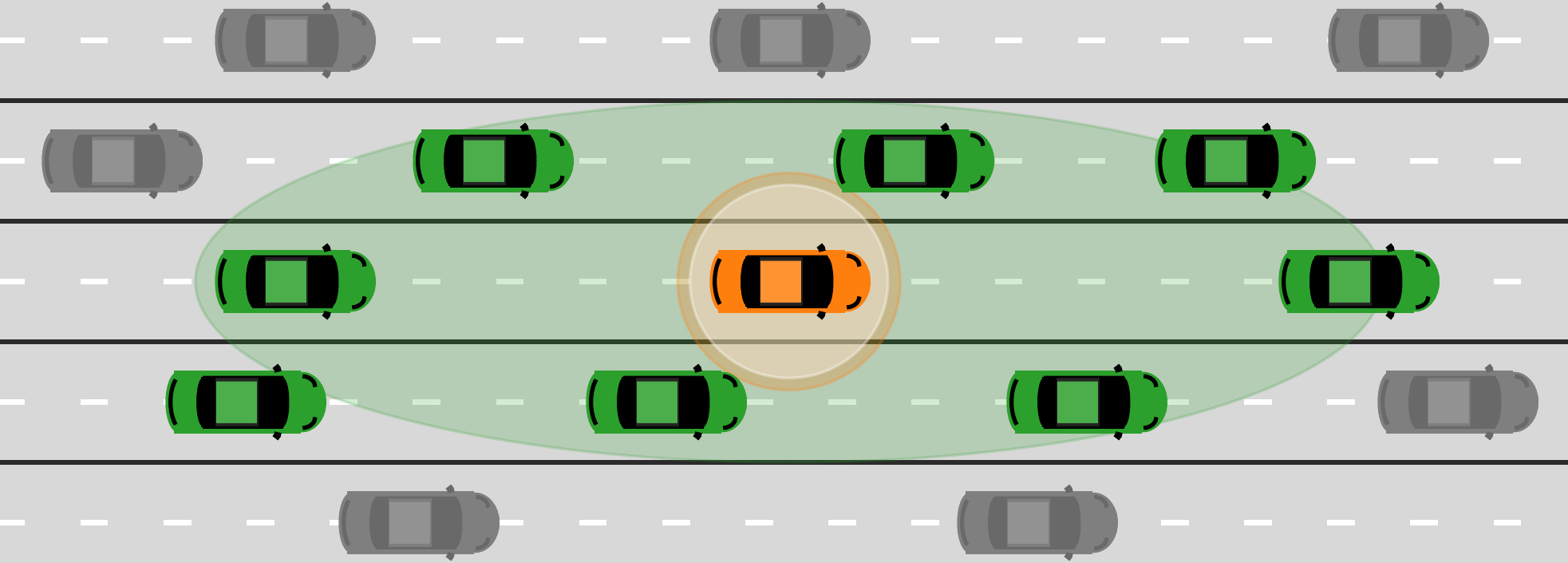}
    \caption{Selection of Surrounding Vehicles.}
    \label{fig:sv_selection}
\end{figure}

\subsubsection{Interaction Features}
With the proven benefit of including information on SVs in the prediction model \cite{deo2018convolutional, hu2018probabilistic, ding2019predicting, messaoud2020attention},
four features are included for every selected SV, which are summarized in \Cref{tab:sv_feat}. 
These are relative lateral and longitudinal positions as well as relative velocities.
The features are calculated according to
\begin{equation}
    \Delta\varphi = \varphi_{\text{SV}} - \varphi_{\text{TV}}, \qquad \forall\varphi\in \{x_l, y_l, v_x, v_y\}.
\end{equation}

The inclusion of information on SVs is the primary contribution of interaction-aware models.
However, there are varying approaches to how such information should be obtained and modeled within a behavior prediction module.
In \cite{deo2018convolutional, messaoud2020attention} a region around the TV is defined and all vehicles that are included within the borders are selected as SVs.
In this work, an approach similar to that in \cite{altche2017lstm, hu2018probabilistic, ding2019predicting} is adopted,
 and so an upper limit of eight SVs are considered as illustrated in \Cref{fig:sv_selection}.
This corresponds to a preceding and following vehicle in the same lane as the TV.
In the neighboring lanes, three vehicles are selected: one immediate neighbor as well as the corresponding preceding and following vehicles of that neighbor.
The vehicles which adhere to the provided selection criteria at the end of the observation are selected as SVs.
For vehicles that are not within $100$ m of the TV, synthetic vehicles are created in their place with the same velocity as the TV.

\begin{table}[!h]
	\vspace{-0.1in}
	\caption{Interaction feature selection}
	\label{tab:sv_feat}
	\centering
	\begin{tabular}{c l c}
		\toprule
		Feature & Description & Unit \\
		\midrule
		$\Delta y$ & Relative lateral position & m\\ 
		$\Delta x$ & Relative longitudinal position & m\\
		$\Delta v_y$ & Relative lateral velocity & m/s \\
		$\Delta v_x$ & Relative longitudinal velocity & m/s  \\
		\bottomrule
	\end{tabular}
\end{table}

\begin{figure}[!t]
	\vspace{-0.2in}
    \centering
    \includegraphics[width=3.5in]{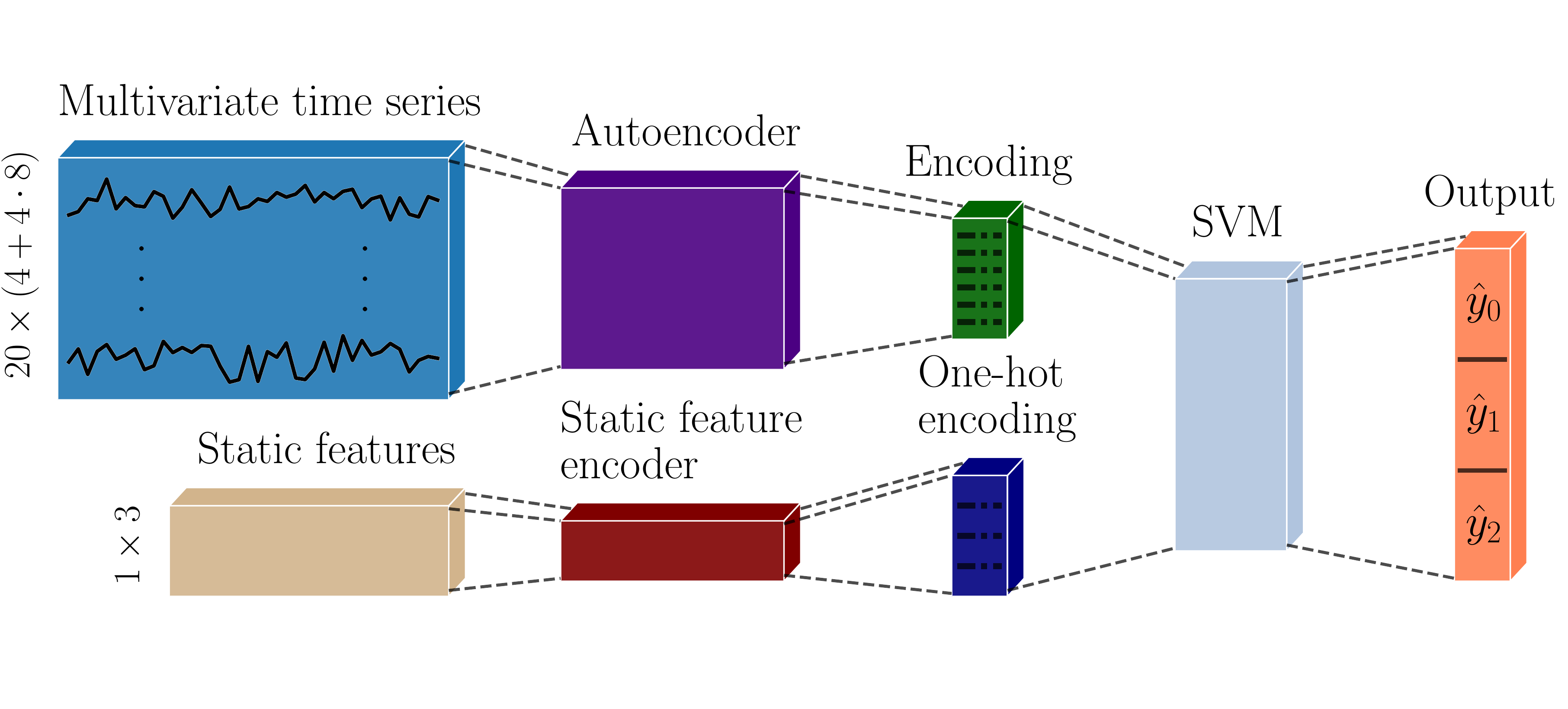}
    \vspace{-0.3in}
    \caption{Overview of model architecture.}
    \label{fig:system_architecture}
\end{figure}

\subsubsection{Static Features}
The static features represent information that is time-invariant.
These may include generic traffic rules, vehicle class, number of lanes, lane type, and so on.
Inclusion of variants of these features are found in some prior works, e.g., \cite{messaoud2020attention, altche2017lstm, phillips2017generalizable}.
Many of these features contain relevant information for the prediction problem, e.g., by imposing dynamic constraints by a vehicle class feature or limiting available actions by a traffic rule--type feature.
Static features are primarily included here for the generalization study and should therefore be
applicable to both data sets. 
Similarly to \cite{messaoud2020attention, altche2017lstm}, vehicle class is included as a feature,
which is available from the data sets (motorcycle (M), automobile (A),
truck (T)). Related to the rule features proposed in \cite{phillips2017generalizable}, 
neighboring lane existence is also included, expressed by
either a 1 (lane exists) or 0 (no lane).  For use in the prediction
model, all static features are encoded using a \emph{one-hot}
scheme (see \Cref{tab:struc_feat}).

\begin{table}[!b]
	\vspace{-0.1in}
	\caption{Static feature selection}
	\label{tab:struc_feat}
	\centering
	\begin{tabular}{c l l}
		\toprule
		Feature & Description & One-hot encoding\\
		\midrule
		$\mathcal{C}$ & Vehicle class & M = $[1,0,0]$, A = $[0,1,0]$, T = $[0,0,1]$\\ 
		$L_{\exists}$ & Left lane exists &  Yes = $[0,1]$, No = $[1,0]$\\
		$R_{\exists}$ & Right lane exists &  Yes = $[0,1]$, No = $[1,0]$\\
		\bottomrule
	\end{tabular}
\end{table}

\subsection{Model Architecture}
\label{sec:sys_arch}
Model development follows a simplistic design principle with focus on
low model complexity. The proposed model, illustrated in \Cref{fig:system_architecture}, consists of an LSTM autoencoder
coupled with a Support Vector Classifier.  The role of the autoencoder
is to act as an automatic feature extractor and provide an efficient
encoding of the input data to the subsequent classifier.

\begin{figure}[!t]
    \centering
    \includegraphics[width=3.3in]{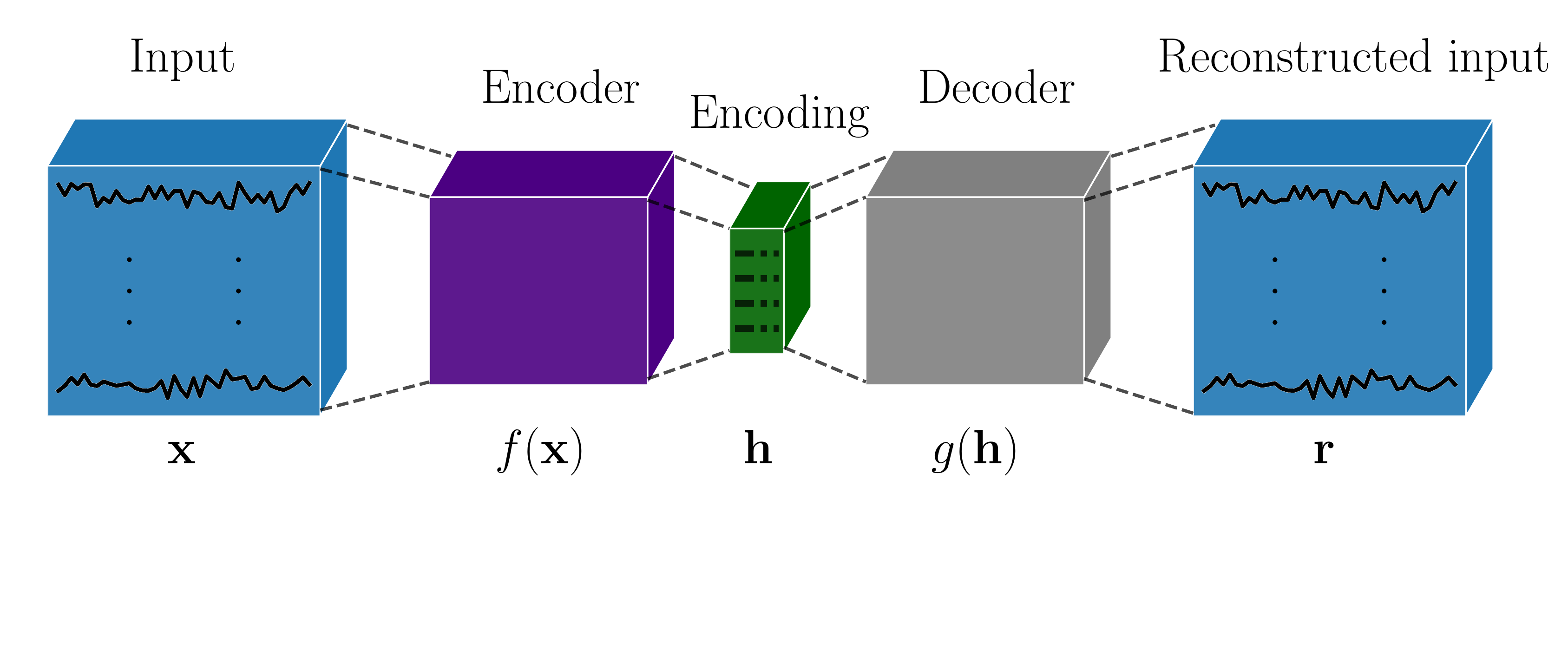}
    \vspace{-0.4in}
    \caption{Autoencoder structure.}
    \label{fig:ae_structure}
\end{figure}

\subsubsection{Autoencoder}

Autoencoders (AEs), illustrated in Fig~\ref{fig:ae_structure}, have been used for
dimensionality reduction or feature scaling
\cite{goodfellow2016deep}.  This is an attractive property for the
considered problem due to the large feature space.  For any learning
problem in general, it is a difficult task to determine which features
in the data set would be most useful.  This is especially true for the
considered application where the inclusion of interaction features
requires the assumption that a predefined selection of SVs would
influence the decision of the TV.  Therefore, assigning the task of
determining feature importance to the model within the learning
process could help the overall classification performance.  Another
interesting property with dimensionality reduction is that it might
offer some noise suppression, a useful feature for the considered data
sets.

With the input data consisting of time-series, it is required that the model should be able to process sequential features.
Recurrent neural networks (RNNs) have historically been used for sequence modeling, and in particular for natural language processing \cite{goodfellow2016deep}.
Although convolutional neural networks is a popular alternative for both time-series classification \cite{fawaz2020inceptiontime} and forecasting \cite{lim2021time}, 
this paper employs an RNN model for sequence encoding.
Specifically, the AE consists of LSTM RNNs where the encoding consists of the final hidden layer of the encoder.
Once the AE model is fit to the training data, only the encoder part is used in the prediction model $\mathcal{M}$.

There are variants of autoencoders that have been proven to be especially useful for classification tasks, e.g., \emph{sparse} autoencoders \cite{makhzani2014k}.
Sparse AEs allow the dimension of the encoding to exceed that of the input but typically only allow a few hidden units to be active at the same time, 
and thus encourage sparsity \cite{goodfellow2016deep}.
However, for the investigated problem, 
dimensionality reduction is more attractive since the time sequence will be compressed.

With the length of the input sequence as well as number of features,
 the main design parameter for the LSTM AE is the \textit{embedding} size that determines the dimension of the reduced encoding.
Given that the embedding size should be less than the flattened input
sequence (feature size is $720$ with a history of $20$ samples for the $36$
features) to achieve dimensionality reduction, 
the sizes considered were $[64, 128, 256, 512]$. 
The embedding size was chosen by iterative evaluation and the
largest size provided the best classifier performance.

\subsubsection{Static Feature Encoder}
The static features are not sequential by nature and are therefore not included in the input of the autoencoder.
Instead, they are first passed through the static feature encoder (SFE) that encodes them using the one-hot scheme.
The static encoding is concatenated with the AE-encoding before passed into the classifier.

\subsubsection{Classifier}
Using the encoding produced by the AE and SFE for classification may
be done using any conventional method, for example with a
fully-connected multilayer perceptron network.  In this work, however,
we use a support vector machine (SVM) model\cite{cortes1995support}, a classical technique
that has been successfully applied to many real-world classification
problems. The use of SVMs is appealing for
any general classification task, attributing their attractiveness to
their theoretical and practical advantages. Finally, an SVM for
classification additionally complements the simplistic modeling approach.

\subsection{Imbalanced Learning}
Real-world data sets are rarely balanced but most often it is the rare occurrences, i.e.,
 the minority class instances we are most interested in detecting or classifying.
In a recent related study by Mirus \textit{et al.} \cite{mirus2020importance}, the authors investigated the inherent importance of a balanced data set for learning a trajectory prediction model.
They found that the choice of data points used in model training and validation has significant impact on the predictive performance.

In this work, two categories of (imbalanced) learning techniques are combined for the learning task, namely \textit{sampling methods} and \textit{ensemble methods}.

Sampling methods are used for class redistribution by direct manipulation of the imbalanced training set \cite{he2013imbalanced} and make up two categories.
\emph{Oversampling} targets the minority class and aims to even out the distribution by aggregating more minority class samples.
\emph{Undersampling} instead targets the majority class by selective removal of corresponding class samples.
In practice, sampling may either be done in a random fashion or in an informed manner \cite{he2013imbalanced}.

Ensemble methods comprise a family of learning techniques that aim to train separate \textit{base learners} for combined use \cite{zhou2012ensemble}.
This includes two major branches of learning techniques known as \textit{boosting} and \textit{bagging} \cite{zhou2012ensemble}.
Conclusively, the goal of both methods are shared: to train several base learners to be combined into a strong learner. 

\subsubsection{Multiclass Balancing Ensemble}
\label{sec:proposed_method_training}
A combination of undersampling and bagging is adopted during training. 
The method is inspired by the \textit{Easy Ensemble} approach proposed in
\cite{liu2008exploratory}. 
Notably, our method is designed for multiclass classification.

Consider a classification problem with one majority class, and $\mu \geq 1$ minority classes.
Let $n_-$ denote the majority class samples and $n_{j+}$ the minority class samples for $j\in\{1,2, \dots, \mu\}$.
Furthermore, let $\mathbb{X}=\mathbb{X_{\minus}} \cup \mathbb{X_+}$ be the training set,
where $\mathbb{X_{\minus}}$ and $\mathbb{X_+} = \bigcup_{j=1}^\mu \mathbb{X}_{j+}$ are the set of samples from the majority and minority classes, respectively.
Furthermore, let $\beta$ denote the number of base learners to train and $\mathcal{T}$ the number of training iterations.
Model training is then performed according to Algorithm~\ref{alg:ensemble}.

\begin{algorithm}[!t]
    \caption{Multiclass Balancing Ensemble}
    \label{alg:ensemble}
    \small
    \begin{algorithmic}[1]
        \FOR{$i=1$ to $\beta$}
        \STATE Randomly sample subset $\mathbb{X}_{i\minus}$ s.t. $\left\lvert\mathbb{X}_{i\minus}\right\rvert=\left\lceil\frac{1}{\mu}\sum_{j=1}^\mu n_{j+}\right\rceil$ 
        \FOR{$k=1$ to $\mathcal{T}$}
        \STATE Train model $\mathcal{M}_i$ on $\mathbb{X}_{i\minus} \cup \ \mathbb{X_+}$
        \ENDFOR
        \ENDFOR
    \end{algorithmic}
	\vspace{-0.01in}
\end{algorithm}

The strength of this method is that it applies the effectiveness of undersampling, without completely removing majority class instances;
instead they are distributed over different \emph{bags}.
Collectively, the base learners $\mathcal{M}_i$ have been trained on a much larger data set than what conventional methods allow.
Once joined, their combined accuracy should be greater than that of the individual base learners.
Here, the aggregation of base learners into an ensemble is done using \emph{soft voting} \cite{zhou2012ensemble}, 
i.e., the combined prediction used for inference is determined by the mean of all outputs of the individual learners.


\section{Evaluation and Results}

\subsection{Evaluation Setup}
Evaluations were performed with the following models:
\begin{itemize}
	\item \textbf{Vanilla CNN} \emph{V-CNN}: A non-interaction-aware single layered convolutional neural network.
	\item \textbf{Social-pooling LSTM} \emph{SP-LSTM}: An LSTM--encoder network with social-pooling adopted from \cite{alahi2016social}.
	\item \textbf{Social-pooling CNN} \emph{SP-CNN}: A CNN with social-pooling adopted from \cite{alahi2016social}.
	\item \textbf{Autoencoder SVM classifier} \emph{AE-SVM}: The proposed model in this work.
	\item \textbf{Autoencoder SVM classifier + SFE}
          \emph{AE-SVM}$^\star$: The proposed model described in \Cref{sec:proposed_method}, equipped with a
          static feature encoder.
\end{itemize}
Unless specified, all networks are coupled with an MLP network for maneuver classification.

\subsubsection{Training and Implementation Details}
The data set is split into train (75\%) and test (25\%) sets by randomly assigning vehicle IDs to each respective set, making sure they are completely disjoint.
For the test set, LKs are discarded in a random fashion such that the set is evenly balanced.
The training-set composition is detailed in Algorithm~\ref{alg:ensemble}.
Additionally, feature scaling is applied through Z-score normalization so that the input data are scaled to have zero mean and unit variance by
\begin{equation}
    \mathbf{z} = \frac{\mathbf{x}-\mu_\mathbf{x}}{s},
\end{equation}
where $\mu_\mathbf{x}$ and $s$ are the mean and biased standard deviation of the training set feature vector, respectively.

A batch size of 256 was used during training, and a \emph{smooth L1} (\emph{Huber})
loss function was used for training the autoencoder.  To mitigate the
exploding gradient problem, clipping \cite{goodfellow2016deep} was
applied during training according to
\begin{equation}
    \mathbf{g} \leftarrow \frac{\mathbf{g}\varepsilon}{\lVert \mathbf{g} \rVert}, \quad \text{if} \ \lVert \mathbf{g} \rVert > \varepsilon,
\end{equation}
where $\mathbf{g}$ is the parameter gradient and $\varepsilon$ is the norm threshold, set to $0.25$.
This condition is checked before every gradient update.
All classifiers were trained using Cross-entropy loss and the optimizer
\textit{AdamW} \cite{loshchilov2017decoupled} was used with a learning rate of $0.0001$.
All implementations were done in PyTorch \cite{paszke2019pytorch}.

\subsubsection{Evaluation Metrics}
\label{sec:eval_metrics}
An approach found in related works is to reformulate the problem as a binary classification task \cite{dang2017time, hu2018probabilistic, ding2019predicting, woo2017lane}.
Adopted by \cite{hu2018probabilistic, ding2019predicting}, 
the authors in \cite{dang2017time} employ the following definitions:
True positive (TP) represents correct prediction of either lane-change left or right. 
False positive (FP) represents mispredicting the LC direction.
False negative (FN) means incorrectly predicting an LC into LK.
Precision, recall, and the $F_1$ score are then commonly calculated as:
\begin{table}[h]
	\centering
	\begin{tabular}{c c c}
		Precision $=\dfrac{\text{TP}}{\text{TP+FP}}$,
		& Recall $=\dfrac{\text{TP}}{\text{TP+FN}}$,
		& $F_1 = 2\dfrac{\text{precision} \cdot \text{recall}}{\text{precision} + \text{recall}}$.\\ [0.5ex] 
	\end{tabular}
\end{table}

Furthermore, the authors in \cite{ding2019predicting} define \emph{critical FNs} to comprise misclassification instances within the TTLC duration of 1.5 s and use them to calculate the recall value. 
For convenient comparison with prior works, these frequency measures are adopted as one set of measures with minor modifications.
\emph{Critical FNs} are used to compute minority class recall, however, a stricter definition of FPs is employed by simply defining them as LK instances predicted as LCs.

To make the evaluation metrics more closely connected to the intention prediction problem in particular, 
we also consider a stricter way to measure classification performance that preserves the formulation as a multiclass classification problem.
Importantly, this approach includes no tailor-made specifications of the frequency measures (TP, FP, FN).
For example, if TP$_{\text{LCL}}$ represents correct prediction of a lane-change left instance, FN$_{\text{LCL}}$ indicates mispredicting the LCL as \emph{either} a LK or LCR regardless of the TTLC.
Preserving the formulation in this way is motivated by its usability in a practical setting:
It is equally important to distinguish between LC direction as it is between LCs and LK instances for the sake of EV local motion planning.
Distinctions between the two performance measures will be highlighted in the text.

\subsection{Results}
\label{sec:results}

\subsubsection{Applying Multiclass Balancing Ensemble}
\label{sec:mcbe_res}
\begin{figure}[!t]
	\centering
	\includegraphics[width=3.2in]{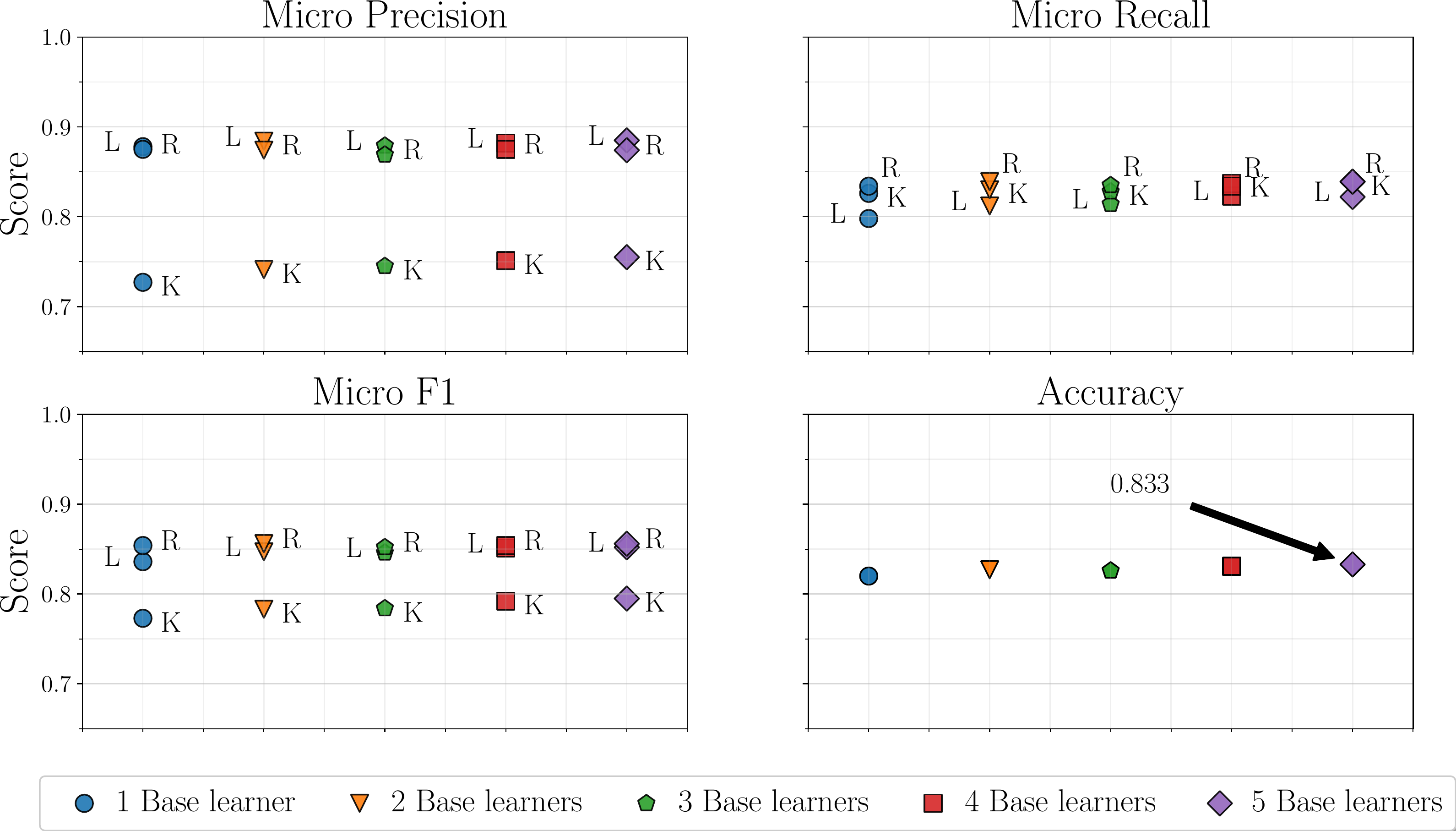}
	\caption{Multiclass classification performance.}
	\label{fig:metrics}
	\vspace{-0.1in}
\end{figure}

The \textit{MCBE} technique outlined in \Cref{sec:proposed_method_training} is evaluated on the full data set (\Cref{tab:data_imbalance}).
The multiclass classification performance of the model is illustrated in
\Cref{fig:metrics} for increasing number of base learners.
The results show that even one additional base learner
for collective inference improves the overall classification
performance.
Precision on LK prediction is seemingly most affected, indicating that there is useful information to be extracted by full use of the set of LK instances.
Although a maximum number of 20 base learners were considered, evaluations showed only minimal performance increase after using 5 base learners, 
indicating that there is a limit to the additional information available in the set of LKs.

Balancing the data set is found to be crucial for classification performance.
For example, running the code provided in \cite{deo2018convolutional} and extracting the intention
prediction module gives recall performance for $(\text{LCL}, \text{LK}, \text{LCR})$ as
$(0.598, 0.996, 0.431)$. 
Here, the class imbalance problem is clearly visible. It must be noted that the main objective of the model in \cite{deo2018convolutional} is trajectory prediction, not intention prediction, but it is a clear illustration of the problem even when using a state-of-the-art method.
Compared with the reported performance in \Cref{fig:metrics}, using our proposed method illustrates significant improvement.

\begin{figure}[ht]
	\centering
	\subfloat[Single random base learner.]{%
		\label{fig:single_base_learner}
		\includegraphics[clip,width=3.3in]{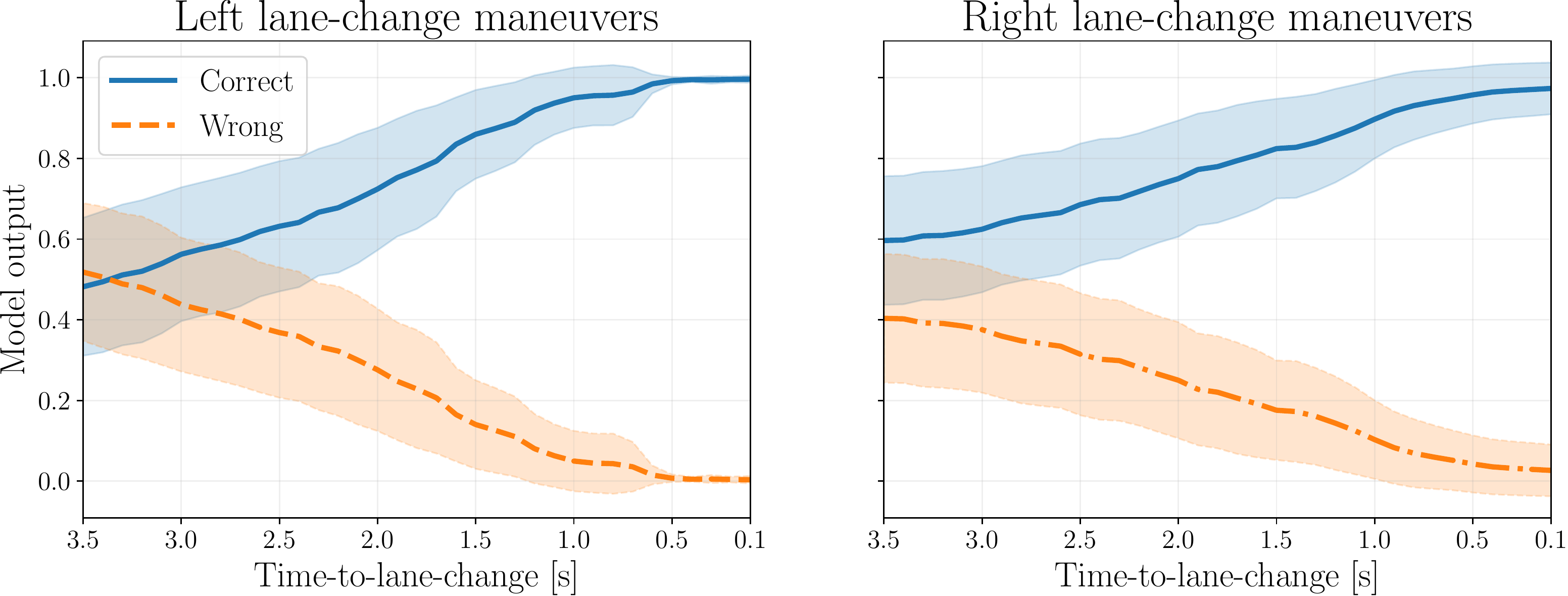}%
	}
	
	\subfloat[Ensemble of 20 base learners.]{%
		\label{fig:ensemble_of_learners}
		\includegraphics[clip,width=3.3in]{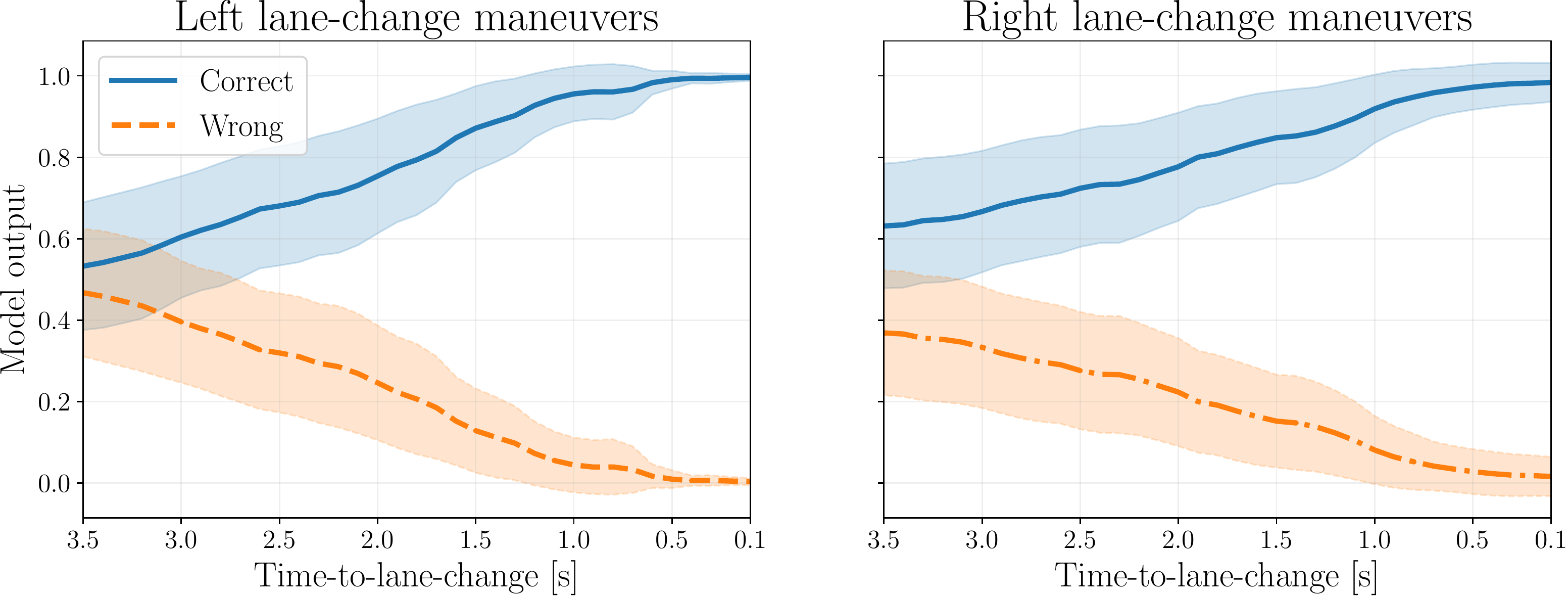}%
	}
	\caption{Mean and standard deviation of predicted lane-change probability.}
	\label{fig:intent_prob}
\end{figure}

The prediction confidence when assessing lane-change instances at decreasing TTLC is presented in \Cref{fig:intent_prob}.
The figure shows how the mean and standard deviation evolves as the TV is approaching the TTLC instant.
Figure \ref{fig:single_base_learner} shows performance of one randomly chosen base learner, 
and \Cref{fig:ensemble_of_learners} shows the collective prediction of an ensemble with all 20 base learners.
The benefit of the \textit{MCBE} technique is perhaps most prominent when studying \Cref{fig:intent_prob}.
Although the variance in the confidence is only slightly affected,
a higher mean can  be observed, resulting in greater separation between the lines.
This has a positive effect on the LC prediction performance, indicating that LCs may be detected at an earlier stage, around half a second earlier than when using one base learner.

To compare the performance with prior works, the modified binary classification measures together with full test set accuracy for an increasing number of base learners in the ensemble are shown in \Cref{tab:full_results}.
The correlation between performance and number of base learners seems to hold for this purpose as well, 
and similarly to what was indicated previously---performance reaches its peak when combining 5 base learners.

\begin{table}[!h]
	\vspace{-0.15in}
	\caption{Performance by adopted metrics}
	\label{tab:full_results}
	\centering
	\begin{tabular}{c l l l l}
		\toprule
		Model & Precision & Recall & $F_1$ & Accuracy\\
		\midrule
		\textit{AE-SVM} & $0.904$ & $0.982$ & $0.941$ & $0.819$\\ 
		\textit{AE-SVM2} & $0.907$ & $0.986$ & $0.945$ & $0.827$\\
		\textit{AE-SVM5} & $0.912$ & $0.991$ & $0.949$ & $0.833$\\
		\textit{AE-SVM10} & $0.912$ & $0.988$ & $0.948$ & $0.833$\\
		\textit{AE-SVM15} & $0.910$ & $0.990$ & $0.948$ & $0.833$\\
		\textit{AE-SVM20} & $0.911$ & $0.991$ & $0.949$ & $0.834$\\
		\bottomrule
	\end{tabular}
\end{table}

The results in \Cref{tab:full_results} report contrasting performance to that in \Cref{fig:metrics}.
It should be noted once again that this evaluation only considers a small number of used FNs, namely those within the 1.5 s TTLC duration.
However, studying \Cref{fig:intent_prob}, it is clear that prediction difficulties arise for TTLC values of 2 s and higher, illustrating the importance to consider performance over the full prediction horizon.

Finally, results are provided for different models using the data selection strategies found in prior works.
Both US-101 and I80 are combined for training and testing, but with a sparse amount of maneuvers,
reflecting a less generous collection or selective removal of LCs during extraction.
The training and test sets are then balanced by removing LK instances in a random fashion until an even class distribution is obtained.

Results are presented using the adopted metrics described in \Cref{sec:eval_metrics} together with the accuracy of the models on the full test set to partly illustrate their multiclass classification performance.
Additionally, the recall score of the LKs that were randomly removed prior to training is included in the results.
Following this convention, the resulting performance of the considered models is presented in \Cref{tab:undersampling_results}.

\begin{table}[!b]
	\vspace{-0.15in}
	\caption{Baseline performance by adopted metrics}
	\label{tab:undersampling_results}
	\centering
	\begin{tabular}{c l l l l l}
		\toprule
		Model & Precision & Recall & $F_1$ & Accuracy & LK Rec.\\
		\midrule
		\textit{V-CNN} & $0.849$ & $0.993$ & $0.915$ & $0.774$ & $0.698$\\ 
		\textit{SP-LSTM} & $0.905$ & $0.987$ & $0.945$ & $0.815$ & $0.826$\\
		\textit{SP-CNN} & $0.881$ & $0.988$ & $0.931$ & $0.785$ & $0.776$\\
		\midrule
		\textit{AE-SVM} & $0.872$ & $0.990$ & $0.927$ & $0.804$ & $0.754$\\
		\textit{AE-SVM}$^\star$ & $0.883$ & $0.991$ & $0.934$ & $0.808$ & $0.777$\\
		\bottomrule
	\end{tabular}
\end{table}
The reported performance in \Cref{tab:undersampling_results},
presents a clear benefit of using interaction-aware models.
Furthermore, regardless of its more simplistic nature compared to some of the other evaluated
models, a single \textit{AE-SVM} model shows competitive performance against its
counterparts.
The results from using \textit{AE-SVM}$^\star$ show that the addition of static features has an overall positive effect on performance
Finally, the results also illustrate a problem with
using a data set that has been undersampled to a large degree.
The recall score on the remaining LKs that have been randomly removed to
balance the data set is well below the reported performance in \Cref{fig:metrics}.
Interestingly, the use of the \textit{MCBE} technique surpasses all models presented here despite being tasked with assessing a much larger data set (cf. \Cref{tab:full_results}).


\subsubsection{Generalization Study}
\label{sec:gen_study}
In a practical application, vehicle behavior predictors should offer similar performance across different traffic scenarios.
Generalization properties are here investigated by training the proposed model on one of the data sets and evaluating it on the other.

Results from training the model on US-101 data and evaluating its performance when tasked with observations from I80 are presented in \Cref{tab:us101vi80}.
Following the trend of previous results, it is clear that the use of \emph{MCBE} is beneficial for overall performance increase.
Similarly to the results presented previously, 
performance reaches its peak when 5 base learners are arranged into an ensemble and so remaining combinations are omitted here for brevity.
Interestingly, including static features within the classification input seems to have a positive effect on the prediction performance.
The most prominent effect is shown through the LK recall score, 
and we hypothesize that including information on existing neighboring lanes is the main benefit here, 
arguably since it limits available actions in the outermost lanes.

\begin{table}[!h]
	\vspace{-0.02in}
	\caption{US-101 model on I80 data multiclass classification performance}
	\label{tab:us101vi80}
	\centering
	\scriptsize
	\setlength\tabcolsep{5.3pt}
	\begin{tabular}{c c c c c c c c}
		\toprule
		Model & \multicolumn{3}{c}{Precision} & \multicolumn{3}{c}{Recall} & Acc.\\ 
		\cmidrule{2-7}
		 & \textit{\scriptsize{LCL}} & \textit{\scriptsize{LK}} & \textit{\scriptsize{LCR}}& \textit{\scriptsize{LCL}} & \textit{\scriptsize{LK}} & \textit{\scriptsize{LCR}} &\\
		\midrule
		\textit{AE-SVM}  & $0.691$ & $0.780$ & $0.700$ & $0.852$ & $0.343$ & $0.926$ & $0.708$\\
		\textit{AE-SVM5} & $0.755$ & $0.769$ & $0.719$ & $0.823$ & $0.474$ & $0.933$ & $0.743$\\
		\midrule
		\textit{AE-SVM}$^\star$ & $0.762$ & $0.705$ & $0.787$ & $0.790$ & $0.581$ & $0.897$ & $0.756$\\
		\textit{AE-SVM5}$^\star$ & $0.775$ & $0.710$ & $0.798$ & $0.792$ & $0.606$ & $0.900$ & $0.765$\\
		\bottomrule
	\end{tabular}
\end{table}

Results from training the model on I80 data and evaluating on US-101 are presented in \Cref{tab:i80vus101}.
The performance in this case shows contrasting trends in comparison with the previous study.
Notably, there is a drop in LC recall for both models.
One probable reason may be explained by the class distribution between the data sets (\Cref{tab:data_imbalance}), 
where the inherent number of LCs differ between data sets---affecting the training and test set composition.
Although the use of \emph{MCBE} still offers positive improvements,
the effect of including static features is ambiguous.
Indicated by the results, the inclusion of static features causes the model to more frequently predict LKs, resulting in lower precision but higher recall.
This is likely attributed to the underlying behavioral differences between the two traffic scenarios.
A non-existing neighboring lane on the less dynamic I80 is indicative of a LK maneuver, which is not necessarily the case on the US-101.

\begin{table}[!h]
	\vspace{-0.08in}
	\caption{I80 model on US-101 data multiclass classification performance}
	\label{tab:i80vus101}
	\centering
	\scriptsize
	\setlength\tabcolsep{5.3pt}
	\begin{tabular}{c c c c c c c c}
		\toprule
		Model & \multicolumn{3}{c}{Precision} & \multicolumn{3}{c}{Recall} & Acc.\\ 
		\cmidrule{2-7}
		& \textit{\scriptsize{LCL}} & \textit{\scriptsize{LK}} & \textit{\scriptsize{LCR}}& \textit{\scriptsize{LCL}} & \textit{\scriptsize{LK}} & \textit{\scriptsize{LCR}} &\\ 
		\midrule
		\textit{AE-SVM}  & $0.762$ & $0.655$ & $0.853$ & $0.793$ & $0.788$ & $0.659$ & $0.744$\\
		\textit{AE-SVM5} & $0.798$ & $0.666$ & $0.840$ & $0.797$ & $0.783$ & $0.695$ & $0.758$\\
		\midrule
		\textit{AE-SVM}$^\star$ &$ 0.882$ & $0.579$ & $0.865$ & $0.683$ & $0.883$ & $0.606 $& $0.725$\\
		\textit{AE-SVM5}$^\star$ & $0.897$ & $0.592$ & $0.874$ & $0.707$ & $0.899$ & $0.606$ & $0.738$\\
		\bottomrule
\end{tabular}
\end{table}


\section{Conclusion}
We have presented an interaction-aware vehicle-intention prediction model consisting of an LSTM autoencoder and SVM classifier.
Model performance is enhanced by using the proposed \emph{multiclass balancing ensemble} method that offers systematic use of the full data set. 
The improved performance by our method demonstrates that there is potentially useful information in commonly disregarded samples---emphasizing the use of methodical techniques for balancing data sets.
A generalization study has been conducted, where performance of the model was evaluated based on its ability to predict maneuvers using observations from unseen traffic scenarios.
It was shown that use of imbalanced learning techniques also carries over to generalization.
Finally, the inclusion of static features within the prediction task showed positive effect on the overall performance.

\bibliographystyle{IEEEtran}
\bibliography{IEEEabrv,references.bib}{}

\begin{thebibliography}{10}
\providecommand{\url}[1]{#1}
\csname url@samestyle\endcsname
\providecommand{\newblock}{\relax}
\providecommand{\bibinfo}[2]{#2}
\providecommand{\BIBentrySTDinterwordspacing}{\spaceskip=0pt\relax}
\providecommand{\BIBentryALTinterwordstretchfactor}{4}
\providecommand{\BIBentryALTinterwordspacing}{\spaceskip=\fontdimen2\font plus
\BIBentryALTinterwordstretchfactor\fontdimen3\font minus
  \fontdimen4\font\relax}
\providecommand{\BIBforeignlanguage}[2]{{%
\expandafter\ifx\csname l@#1\endcsname\relax
\typeout{** WARNING: IEEEtran.bst: No hyphenation pattern has been}%
\typeout{** loaded for the language `#1'. Using the pattern for}%
\typeout{** the default language instead.}%
\else
\language=\csname l@#1\endcsname
\fi
#2}}
\providecommand{\BIBdecl}{\relax}
\BIBdecl

\bibitem{lefevre2014survey}
S.~Lef{\`e}vre, D.~Vasquez, and C.~Laugier, ``A survey on motion prediction and
  risk assessment for intelligent vehicles,'' \emph{ROBOMECH journal}, vol.~1,
  no.~1, pp. 1--14, 2014.

\bibitem{mozaffari2020deep}
S.~Mozaffari, O.~Y. Al-Jarrah, M.~Dianati, P.~Jennings, and A.~Mouzakitis,
  ``Deep learning-based vehicle behavior prediction for autonomous driving
  applications: A review,'' \emph{IEEE Transactions on Intelligent
  Transportation Systems}, 2020.

\bibitem{deo2018convolutional}
N.~Deo and M.~M. Trivedi, ``Convolutional social pooling for vehicle trajectory
  prediction,'' in \emph{Proceedings of the IEEE Conference on Computer Vision
  and Pattern Recognition Workshops}, 2018, pp. 1468--1476.

\bibitem{messaoud2020attention}
K.~Messaoud, I.~Yahiaoui, A.~Verroust-Blondet, and F.~Nashashibi, ``Attention
  based vehicle trajectory prediction,'' \emph{IEEE Transactions on Intelligent
  Vehicles}, vol.~6, no.~1, pp. 175--185, 2021.

\bibitem{hu2018probabilistic}
Y.~Hu, W.~Zhan, and M.~Tomizuka, ``Probabilistic prediction of vehicle semantic
  intention and motion,'' in \emph{IEEE Intelligent Vehicles Symposium (IV)},
  2018, pp. 307--313.

\bibitem{ding2019predicting}
W.~Ding, J.~Chen, and S.~Shen, ``Predicting vehicle behaviors over an extended
  horizon using behavior interaction network,'' in \emph{IEEE International
  Conference on Robotics and Automation (ICRA)}, 2019, pp. 8634--8640.

\bibitem{he2013imbalanced}
H.~He and Y.~Ma, \emph{Imbalanced learning: Foundations, algorithms, and
  applications}.\hskip 1em plus 0.5em minus 0.4em\relax Hoboken, New Jersey:
  John Wiley \& Sons, 2013.

\bibitem{kumar2013learning}
P.~Kumar, M.~Perrollaz, S.~Lefevre, and C.~Laugier, ``Learning-based approach
  for online lane change intention prediction,'' in \emph{IEEE Intelligent
  Vehicles Symposium (IV)}, 2013, pp. 797--802.

\bibitem{woo2017lane}
H.~Woo, Y.~Ji, H.~Kono, Y.~Tamura, Y.~Kuroda, T.~Sugano, Y.~Yamamoto,
  A.~Yamashita, and H.~Asama, ``Lane-change detection based on
  vehicle-trajectory prediction,'' \emph{IEEE Robotics and Automation Letters},
  vol.~2, no.~2, pp. 1109--1116, 2017.

\bibitem{streubel2014prediction}
T.~Streubel and K.~H. Hoffmann, ``Prediction of driver intended path at
  intersections,'' in \emph{IEEE Intelligent Vehicles Symposium (IV)}, 2014,
  pp. 134--139.

\bibitem{deo2018would}
N.~Deo, A.~Rangesh, and M.~M. Trivedi, ``How would surround vehicles move? {A}
  unified framework for maneuver classification and motion prediction,''
  \emph{IEEE Transactions on Intelligent Vehicles}, vol.~3, no.~2, pp.
  129--140, 2018.

\bibitem{yoon2016multilayer}
S.~Yoon and D.~Kum, ``The multilayer perceptron approach to lateral motion
  prediction of surrounding vehicles for autonomous vehicles,'' in \emph{IEEE
  Intelligent Vehicles Symposium (IV)}, 2016, pp. 1307--1312.

\bibitem{phillips2017generalizable}
D.~J. {Phillips}, T.~A. {Wheeler}, and M.~J. {Kochenderfer}, ``Generalizable
  intention prediction of human drivers at intersections,'' in \emph{IEEE
  Intelligent Vehicles Symposium (IV)}, 2017, pp. 1665--1670.

\bibitem{zyner2017long}
A.~Zyner, S.~Worrall, J.~Ward, and E.~Nebot, ``Long short term memory for
  driver intent prediction,'' in \emph{IEEE Intelligent Vehicles Symposium
  (IV)}, 2017, pp. 1484--1489.

\bibitem{dang2017time}
H.~Q. {Dang}, J.~{Fürnkranz}, A.~{Biedermann}, and M.~{Hoepfl},
  ``Time-to-lane-change prediction with deep learning,'' in \emph{IEEE 20th
  International Conference on Intelligent Transportation Systems (ITSC)}, 2017.

\bibitem{messaoud2020trajectory}
K.~{Messaoud}, N.~{Deo}, M.~M. {Trivedi}, and F.~{Nashashibi}, ``{Trajectory
  Prediction for Autonomous Driving based on Multi-Head Attention with Joint
  Agent-Map Representation},'' \emph{arXiv:2005.02545 [cs]}, May 2020.

\bibitem{zhao2019multi}
T.~Zhao, Y.~Xu, M.~Monfort, W.~Choi, C.~Baker, Y.~Zhao, Y.~Wang, and Y.~N. Wu,
  ``Multi-agent tensor fusion for contextual trajectory prediction,'' in
  \emph{IEEE/CVF Conference on Computer Vision and Pattern Recognition (CVPR)},
  2019, pp. 12\,118--12\,126.

\bibitem{lee2017convolution}
D.~Lee, Y.~P. Kwon, S.~McMains, and J.~K. Hedrick, ``Convolution neural
  network-based lane change intention prediction of surrounding vehicles for
  {ACC},'' in \emph{IEEE 20th International Conference on Intelligent
  Transportation Systems (ITSC)}, 2017.

\bibitem{casas2018intentnet}
S.~Casas, W.~Luo, and R.~Urtasun, ``Intentnet: Learning to predict intention
  from raw sensor data,'' in \emph{Conference on Robot Learning}.\hskip 1em
  plus 0.5em minus 0.4em\relax PMLR, 2018, pp. 947--956.

\bibitem{cui2019multimodal}
H.~Cui, V.~Radosavljevic, F.-C. Chou, T.-H. Lin, T.~Nguyen, T.-K. Huang,
  J.~Schneider, and N.~Djuric, ``Multimodal trajectory predictions for
  autonomous driving using deep convolutional networks,'' in \emph{IEEE
  International Conference on Robotics and Automation (ICRA)}, 2019, pp.
  2090--2096.

\bibitem{altche2017lstm}
F.~Altch{\'e} and A.~de~La~Fortelle, ``An {LSTM} network for highway trajectory
  prediction,'' in \emph{IEEE 20th International Conference on Intelligent
  Transportation Systems (ITSC)}, 2017, pp. 353--359.

\bibitem{zyner2019naturalistic}
A.~Zyner, S.~Worrall, and E.~Nebot, ``Naturalistic driver intention and path
  prediction using recurrent neural networks,'' \emph{IEEE Transactions on
  Intelligent Transportation Systems}, vol.~21, no.~4, pp. 1584--1594, 2019.

\bibitem{alahi2016social}
A.~Alahi, K.~Goel, V.~Ramanathan, A.~Robicquet, L.~Fei-Fei, and S.~Savarese,
  ``Social {LSTM}: Human trajectory prediction in crowded spaces,'' in
  \emph{Proceedings of the IEEE Conference on Computer Vision and Pattern
  Recognition}, 2016, pp. 961--971.

\bibitem{bahdanau2016neural}
D.~{Bahdanau}, K.~{Cho}, and Y.~{Bengio}, ``{Neural Machine Translation by
  Jointly Learning to Align and Translate},'' \emph{arXiv:1409.0473 [cs]}, Sep.
  2016.

\bibitem{xin2018intention}
L.~Xin, P.~Wang, C.-Y. Chan, J.~Chen, S.~E. Li, and B.~Cheng, ``Intention-aware
  long horizon trajectory prediction of surrounding vehicles using dual {LSTM}
  networks,'' in \emph{IEEE 21st International Conference on Intelligent
  Transportation Systems (ITSC)}, 2018, pp. 1441--1446.

\bibitem{colyar2007us101}
J.~Colyar and J.~Halkias, ``{US} {Highway} 101 {Dataset}. {Federal Highway
  Administration Research and Technology fact sheet.} {Publication} number:
  {FHWA}-{HRT}-07-030,'' \emph{Tech. Rep.}, 2007,
  \url{http://doi.org/10.21949/1504477}.

\bibitem{halkias2006i80}
J.~Halkias and J.~Colyar, ``{Interstate} 80 {Freeway Dataset}. {Federal Highway
  Administration Research and Technology fact sheet.} {Publication} number:
  {FHWA}-{HRT}-06-137,'' \emph{Tech. Rep.}, 2006,
  \url{http://doi.org/10.21949/1504477}.

\bibitem{montanino2013making}
M.~Montanino and V.~Punzo, ``Making {NGSIM} data usable for studies on traffic
  flow theory: Multistep method for vehicle trajectory reconstruction,''
  \emph{Transportation Research Record}, vol. 2390, no.~1, pp. 99--111, 2013.

\bibitem{goodfellow2016deep}
I.~Goodfellow, Y.~Bengio, A.~Courville, and Y.~Bengio, \emph{Deep
  learning}.\hskip 1em plus 0.5em minus 0.4em\relax MIT press, Cambridge, 2016.

\bibitem{fawaz2020inceptiontime}
H.~I. Fawaz, B.~Lucas, G.~Forestier, C.~Pelletier, D.~F. Schmidt, J.~Weber,
  G.~I. Webb, L.~Idoumghar, P.-A. Muller, and F.~Petitjean, ``{Inceptiontime:
  Finding Alexnet for Time Series Classification},'' \emph{Data Mining and
  Knowledge Discovery}, vol.~34, no.~6, pp. 1936--1962, 2020.

\bibitem{lim2021time}
B.~Lim and S.~Zohren, ``Time-series forecasting with deep learning: {A
  survey},'' \emph{Philosophical Transactions of the Royal Society A}, vol.
  379, no. 2194, p. 20200209, 2021.

\bibitem{makhzani2014k}
A.~{Makhzani} and B.~{Frey}, ``{k-Sparse Autoencoders},'' \emph{arXiv:1312.5663
  [cs]}, Mar. 2014.

\bibitem{cortes1995support}
C.~Cortes and V.~Vapnik, ``Support-vector networks,'' \emph{Machine Learning},
  vol.~20, no.~3, pp. 273--297, 1995.

\bibitem{mirus2020importance}
F.~Mirus, T.~C. Stewart, and J.~Conradt, ``The importance of balanced data
  sets: Analyzing a vehicle trajectory prediction model based on neural
  networks and distributed representations,'' in \emph{International Joint
  Conference on Neural Networks (IJCNN)}, 2020.

\bibitem{zhou2012ensemble}
Z.-H. Zhou, \emph{Ensemble methods: Foundations and algorithms}.\hskip 1em plus
  0.5em minus 0.4em\relax Boca Raton, Florida: Chapman \& Hall/CRC, 2012.

\bibitem{liu2008exploratory}
X.~{Liu}, J.~{Wu}, and Z.~{Zhou}, ``Exploratory undersampling for
  class-imbalance learning,'' \emph{IEEE Transactions on Systems, Man, and
  Cybernetics, Part B (Cybernetics)}, vol.~39, no.~2, pp. 539--550, 2009.

\bibitem{loshchilov2017decoupled}
I.~{Loshchilov} and F.~{Hutter}, ``{Decoupled Weight Decay Regularization},''
  \emph{arXiv:1711.05101 [cs]}, Jan. 2019.

\bibitem{paszke2019pytorch}
A.~{Paszke}, S.~{Gross}, F.~{Massa}, A.~{Lerer}, J.~{Bradbury}, G.~{Chanan},
  T.~{Killeen}, Z.~{Lin}, N.~{Gimelshein}, L.~{Antiga}, A.~{Desmaison},
  A.~{K{\"o}pf}, E.~{Yang}, Z.~{DeVito}, M.~{Raison}, A.~{Tejani},
  S.~{Chilamkurthy}, B.~{Steiner}, L.~{Fang}, J.~{Bai}, and S.~{Chintala},
  ``{PyTorch: An Imperative Style, High-Performance Deep Learning Library},''
  \emph{arXiv:1912.01703 [cs]}, Dec. 2019.

\end{thebibliography}

\end{document}